\documentclass[10pt]{article}

\usepackage[preprint]{tmlr}

\usepackage{titleblock}

\renewcommand{\headrulewidth}{0pt}

\renewenvironment{abstract}{\vskip.075in\noindent\ignorespaces}{\par\vskip 1ex}

\usepackage{amsmath,amsfonts,bm}

\def\eqref#1{equation~\ref{#1}}

\def\1{\bm{1}}

\DeclareMathAlphabet{\mathsfit}{\encodingdefault}{\sfdefault}{m}{sl}
\SetMathAlphabet{\mathsfit}{bold}{\encodingdefault}{\sfdefault}{bx}{n}

\usepackage{url}
\usepackage{booktabs}
\usepackage{colortbl}
\usepackage{graphicx}
\usepackage{xcolor}
\definecolor{maintext}{HTML}{1F1F1F}
\AtBeginDocument{\color{maintext}}
\definecolor{refblue}{HTML}{376cd5}
\definecolor{tablerule}{HTML}{e4e4e4}
\arrayrulecolor{tablerule}
\usepackage[colorlinks=true,linkcolor=refblue,citecolor=refblue,urlcolor=refblue]{hyperref}
\usepackage{cleveref}
\usepackage{svg}

\definecolor{clrVI}{HTML}{0F52BA}
\definecolor{clrMO}{HTML}{2A9D8F}
\definecolor{clrRM}{HTML}{E76F51}
\definecolor{clrAA}{HTML}{606C38}
\definecolor{clrAR}{HTML}{7B2D8B}
\definecolor{clrVP}{HTML}{E63946}
\definecolor{clrAS}{HTML}{B5838D}

\newcommand{\inR}{\in \mathbb{R}}

\title{PRAGMA: Revolut Foundation Model}

\author{
Maxim Ostroukhov\textsuperscript{1}\qquad Ruslan Mikhailov\textsuperscript{1}\qquad Vladimir Iashin\textsuperscript{1}\qquad Artem Sokolov\textsuperscript{1}\\[3pt]
Andrei Akshonov\textsuperscript{1}\qquad Vitaly Protasov\textsuperscript{1}\qquad Andrey Goncharov\textsuperscript{1}\qquad Dmitrii Beloborodov\textsuperscript{1}\\[3pt]
Vince Mullin\textsuperscript{2}\qquad Roman Y. Enzmann\textsuperscript{2}\qquad Georgios Kolovos\textsuperscript{2}\qquad Jason Renders\textsuperscript{2}\\[3pt]
Pavel Nesterov\textsuperscript{1}\qquad Anton Repushko\textsuperscript{1}\\[8pt]
{\small\itshape
\textsuperscript{1}Revolut Research\quad\quad\textsuperscript{2}NVIDIA}
}

\begin{document}

\begin{titleblock}
\maketitle

\begin{abstract}
Modern financial systems generate vast quantities of transactional and event-level data that encode rich economic signals.
This paper presents PRAGMA, a family of foundation models for banking event sequences.
Our approach pre-trains a Transformer-based architecture with masked modelling on a large-scale, heterogeneous banking event corpus using a self-supervised objective tailored to the discrete, variable-length nature of financial records.
The resulting model supports a wide range of downstream tasks such as credit scoring, fraud detection, and lifetime value prediction: strong performance can be achieved by training a simple linear model on top of the extracted embeddings and can be further improved with lightweight fine-tuning.
Through extensive evaluation on downstream tasks, we demonstrate that PRAGMA achieves superior performance across multiple domains directly from raw event sequences, providing a general-purpose representation layer for financial applications.\\

\textbf{Disclaimer}: We report only relative improvements, as absolute metrics are commercially sensitive.\\
All examples are synthetic and not from real production data.

\end{abstract}
\end{titleblock}

\begin{figure}[h]
    \vspace{-3em}
    \centering
    \includegraphics[width=1\linewidth]{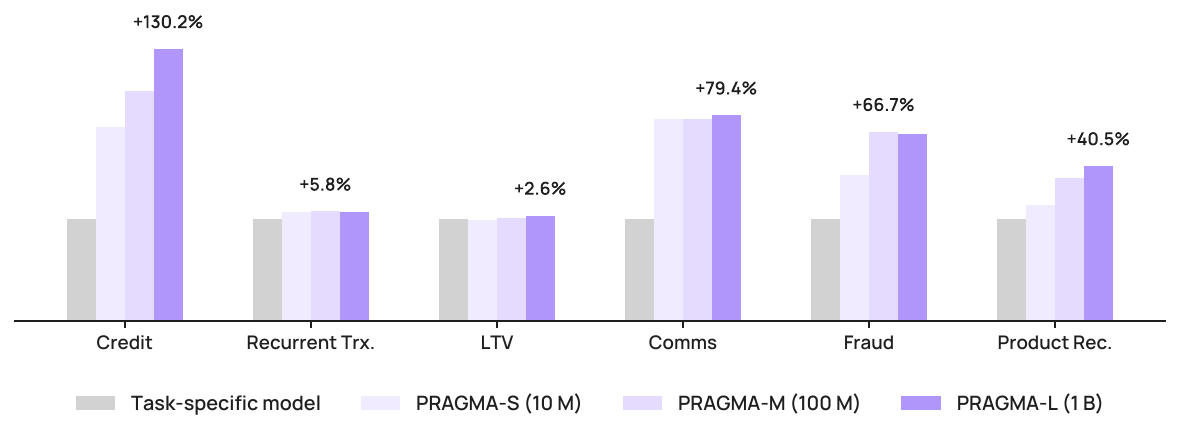}
    \caption{
        A single architecture from 10M to 1B parameters that outperforms task-specific models across~tasks.
    }
    \label{fig:teaser_results}
\end{figure}

\section{Introduction}
\label{sec:introduction}
Foundation models are general-purpose models trained at scale on broad data distributions and subsequently adapted to a wide variety of downstream tasks~\citep{bommasani2021foundation}.
While such models have transformed natural language processing~\citep{devlin2019bert,brown2020language} and computer vision~\citep{kirillov2023segment,caron2021emerging}, their application to multi-source banking user histories remains comparatively underexplored.
Modern banks and fintechs accumulate large volumes of data: event streams spanning multiple event sources such as card and transfer transactions, in-app navigation, and customer communications, alongside static generalised profile state such as account tenure and plan.
These event streams encode signals relevant to risk management, product analytics, and operations, but they are difficult to model efficiently with off-the-shelf language-model tokenisation and architectures.
While serialising structured records as text and feeding them to a standard Transformer is a viable baseline, it inflates sequence lengths considerably because every field name and delimiter becomes several subword tokens.
Moreover, numerical values are split into digit fragments that discard magnitude and ordering, both of which are critical for financial reasoning.
Together, these limitations make naive text serialisation impractical for the long, heterogeneous user histories common in banking.

Multi-source banking user histories differ from text in three ways.
First, each event is a variable-length record with mixed categorical, numerical, and free-text fields.
Second, histories are long-tailed in length and irregular in time, with strong daily and weekly cycles.
Third, practical deployments must operate under strict privacy and regulatory constraints, which limit what can be reported and which features can be used for certain decisions.
Because no single off-the-shelf architecture handles all three challenges simultaneously, practitioners default to building task-specific pipelines with extensive feature engineering, making it hard to share statistical strength across domains and products.

Prior work addresses isolated slices of this problem.
Tabular Transformers such as TabTransformer and FT-Transformer~\citep{huang2020tabtransformer,gorishniy2021revisiting} model fixed-schema rows, while sequential recommender models such as SASRec and BERT4Rec~\citep{kang2018self,sun2019bert4rec} operate on item-like interaction histories.
Financial foundation models have largely focused on text or generic time-series tokenisation~\citep{yang2020finbert,wu2023bloomberggpt,yang2023fingpt,jin2024timellm,ansari2024chronos}, while transaction models such as nuFormer and TransactionGPT~\citep{braithwaite2025your,dou2025transactiongpt} move closer to our setting.
nuFormer pre-trains a decoder on transaction sequences and adds a tabular fusion network during downstream fine-tuning, while TransactionGPT autoregressively models payment transaction histories.
PRAGMA instead uses a dedicated profile-state encoder to learn task-agnostic profile representations jointly with heterogeneous banking events during self-supervised backbone pre-training, then transfers the same backbone across six discriminative tasks.

In this paper, we present PRAGMA, a family of encoder-style foundation models for multi-source banking user histories.
PRAGMA is pre-trained with masked modelling on a large-scale corpus of user histories that combines multi-source events with static profile state~(\S\ref{sec:data}).
To handle heterogeneity, we apply a key--value--time tokenisation scheme with type-specific value encoding for numerical, categorical, and textual fields~(\S\ref{sec:tokenisation}).
The resulting backbone uses two encoder branches for profile state and events whose outputs are fused by a history encoder~(\S\ref{sec:architecture}).

We choose an encoder-only, bidirectional design because our primary goal is transferable representations for discriminative financial tasks, rather than open-ended generation.
Masked modelling enables each token to attend to both past and future context~\citep{devlin2019bert}, which is particularly useful when reconstructing partially observed event records and learning record-level representations from complete histories.
After pre-training, PRAGMA can be adapted efficiently in two complementary ways~(\S\ref{sec:protocol}).
In the \emph{embedding probe} setting, we freeze the backbone and train a lightweight head on top of the extracted embeddings.
In the \emph{LoRA fine-tuning} setting, we apply Low-Rank Adaptation~(LoRA)~\citep{hu2022lora} to update only a small fraction of parameters, enabling fast specialisation while keeping most of the backbone shared across tasks.

We evaluate PRAGMA on a suite of internal downstream benchmarks spanning credit scoring, fraud detection, communication engagement, recurrent transaction detection, lifetime value prediction, and more~(\S\ref{sec:downstreams}).
Across evaluated domains, PRAGMA consistently outperforms strong task-specific baselines while reducing the need for hand-crafted features (Figure~\ref{fig:teaser_results}).
We further describe the engineering choices required to train PRAGMA efficiently on long and highly variable user histories, including sequence packing and dynamic batching~(\S\ref{sec:train_infra}).

Our contributions are as follows:
\begin{itemize}
    \item We introduce PRAGMA, a family of encoder-style foundation models for multi-source banking user histories that scales from 10\,M to 1\,B parameters and is, to our knowledge, the largest published encoder backbone for consumer banking event sequences. The architecture combines a key--value--time tokenisation scheme with a two-branch design in which profile-state and event encoders feed a history encoder for heterogeneous financial records.
    \item We describe an efficient pre-training recipe for long and irregular banking user histories based on masked modelling, sequence packing, and dynamic batching, and show that LoRA fine-tuning of a pre-trained backbone consistently matches or outperforms full training from scratch.
    \item We evaluate a single pre-trained backbone across six diverse downstream tasks: credit scoring, fraud detection, lifetime value, communication engagement, recurrent transaction detection, and product recommendation.
    Across these tasks, PRAGMA consistently outperforms strong task-specific baselines while reducing the need for hand-crafted features.
\end{itemize}

\section{Pre-training}
\label{sec:pretraining-main}
\subsection{Dataset}
\label{sec:data}

\begin{figure}
    \centering
    \includegraphics[width=1\linewidth]{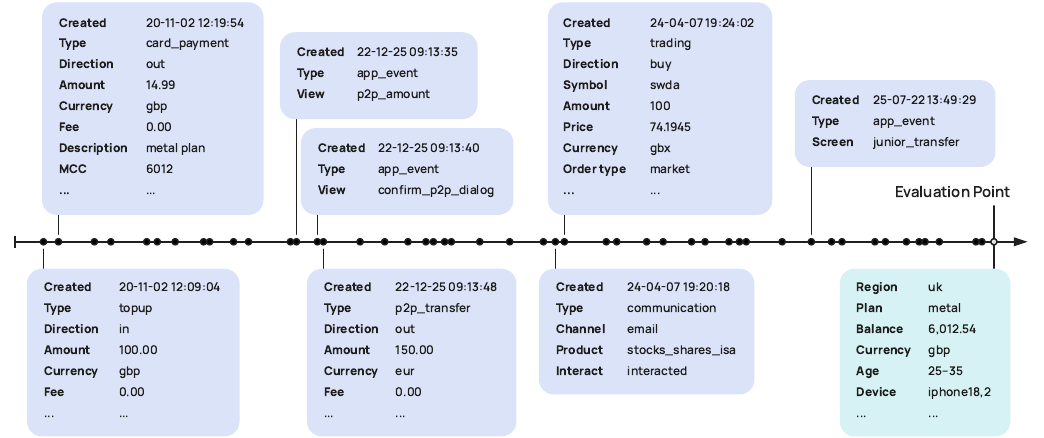}
    \caption{
        \textbf{Event timeline overview}.
        After account creation, users generate a sequence of platform interactions over time, spanning transactions, in-app navigation, and communications.
        We aggregate the event history up until a designated evaluation point.
        Alongside these sequential events, we capture contextual attributes that describe the record's state at that point, e.g., membership plan or service region.
        Both events and attributes share a uniform representation: a timestamp and a set of key--value pairs (e.g., \texttt{Type:} \texttt{card\_payment}, \texttt{Channel:} \texttt{email}).
        All values shown are synthetic; the figure is for illustration purposes only.
    }
    \label{fig:timeline}
\end{figure}

Our goal is to build a foundation model that encodes diverse event-level signals and transfers across a wide range of downstream tasks.
Our dataset is structured at the record level, where each observation represents a pseudonymised event history associated with an evaluation point.
As shown in Figure~\ref{fig:timeline}, we consider an event history alongside contextual attributes.
This approach enables the model to account for both sequential patterns and time-invariant features like account currency.

All data used in this work is fully anonymised and contains no personally identifiable information.
We construct our pre-training dataset from 26\,M user records spanning 111 countries, accumulating 24\,B events that total 207\,B tokens.

\subsubsection{Event History}
Standard platform usage generates event streams across various services, e.g., account funding, payments, in-app navigation, or service communications.
These aggregated event histories capture population-level patterns that support a range of analytical and predictive tasks.
An event is defined by a created timestamp and a set of key--value pairs, e.g., \texttt{Direction:} \texttt{out}.
We fetch events from broad source types that can be loosely grouped into transactions, app, trading, and communication, which were selected for their high expected impact on downstream tasks.
Event schemas are specific to their source type and incorporate distinct sets of keys, e.g., the \texttt{Symbol} key is unique to trading events.
Beyond anonymisation, de-identification, and standard eligibility criteria, no additional statistical filtering or pre-processing, such as outlier removal or vocabulary pruning, is applied to the event streams to ensure that the model captures the full heterogeneity found in production.

\subsubsection{Profile State}
In addition to the event history, we incorporate general contextual attributes such as balance quantile, plan, insurance state, and service region.
These attributes provide useful context that is otherwise missing from the event history alone.
Profile state is a set of descriptive key--value pairs in an event-like format, e.g., \texttt{Plan:} \texttt{metal}, timestamped at the designated evaluation point (or the cut-off date during pre-training).

High-activity users often generate tens of thousands of interactions, exceeding computational bounds; we address this via truncation to a fixed context window~(\S\ref{sec:training}).
However, truncation risks discarding early historical milestones that carry useful signal, such as account age.
We therefore augment profile state with \emph{life-long events}, key--value pairs that, unlike regular profile attributes, each carry an individual timestamp recording a first occurrence, e.g., \texttt{Lifelong:} \texttt{first\_topup} at \texttt{20-11-02 12:09:04}.
This timestamp is then used to compute the temporal distance to the evaluation point, enabling the model to encode the timing of historical milestones.

\subsubsection{Pre-training Time Range}
Developing a robust and generalisable model requires a delicate balance between maximising historical coverage and maintaining data relevance.
Accordingly, determining the optimal temporal range for pre-training involves navigating several trade-offs between event diversity, distribution shift, and computational efficiency.

First, simply including every event from the full available dataset is often impractical and sub-optimal.
Older events may reflect historical patterns, product features, or system dynamics that are no longer relevant at inference time.
Such discrepancies create a distribution mismatch that can degrade performance, as the model may struggle to generalise from obsolete historical examples to the evolving behaviours present in deployment.
Additionally, the inclusion of highly heterogeneous events from long time spans can make the pre-training task harder and slow down model convergence.
Second, downstream applications may require making predictions on events that took place within temporal ranges either much earlier or much later than those used for pre-training.
If the model is not exposed to sufficient diversity in both recent and less-common historical patterns, the performance on these out-of-distribution inputs may suffer.
Finally, Transformer architectures have a limited effective context span, determined both by model design and hardware constraints.

With these considerations in mind, we select a temporal range of 25 months from 2023 to 2025 for pre-training, balancing comprehensive event coverage, recency, distribution consistency, and tractable sequence modelling.

\subsection{Tokenisation}
\label{sec:tokenisation}

Unlike standard LLMs that treat everything as text, a financial foundation model needs to preserve the structural nature and heterogeneity of tabular data.
We address this challenge by implementing a disentangled embedding space of input tokens.

\begin{figure}
    \centering
    \includegraphics[width=1\linewidth]{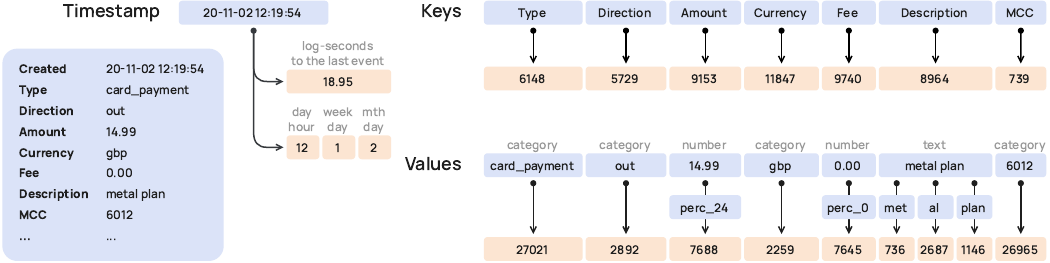}
    \caption{
        \textbf{Tokenisation overview}.
        A raw event record is decomposed into a temporal coordinate, semantic types (keys), and values.
        Keys are always represented by one token, while values use type-specific tokenisation: numerical values are bucketised by percentile, categorical values map to a single token, and textual values are split into subword tokens.
        Some keys therefore expand to multiple value tokens, e.g., \texttt{Description} $\rightarrow$ \texttt{met}, \texttt{al}, \texttt{plan}.
        Time is encoded both as log-seconds to the last event and as calendar and time features derived from the timestamp.
        Profile state is encoded similarly to an event record.
    }
    \label{fig:tokenisation}
\end{figure}

As shown in Figure~\ref{fig:tokenisation}, we represent each data point by three components: a semantic type (key), a value, and a temporal coordinate.
For instance, \texttt{Channel:} \texttt{email} at \texttt{24-04-07 19:20:18} maps to a key, a value, and a temporal coordinate, respectively.
This ensures that the model distinguishes between the meaning of a field and its value, while also encoding event chronology.
Next, we present how the three are tokenised.

\paragraph*{Semantic Type (Key).}
The semantic type embedding enables the model to learn the meaning of a field and to contextualise the value it holds.
We tokenise each semantic type (key) as a single token, and both event and profile state semantic types are encoded in a similar way.
This results in a vocabulary of $\sim$60 tokens.

\paragraph*{Value.}
We cover the diversity of values with three value types: \emph{numerical}, \emph{categorical}, and \emph{textual}.
Numerical values are mapped to percentile buckets whose boundaries are learned from training data.
We add an extra bucket for zero and allocate one token per bucket.
The distinction between categorical and textual values is determined by cardinality thresholding: string fields whose number of unique values falls below a predefined threshold are treated as categorical, while higher-cardinality fields are treated as textual.
Categorical values are manually selected from all text fields to prevent splitting common values, such as merchant category codes (MCC), into multiple tokens, and are represented as a single token as well.
For textual fields, values are tokenised with a BPE-style subword tokeniser~\citep{sennrich2016neural}
with a reserved \texttt{[UNK]} token for rare unseen fragments.
In total, value tokens use a vocabulary of ${\sim}$28\,k tokens.

\paragraph*{Temporal Information.}
We encode time in two ways.
First, we compute the elapsed time since the most recent event, measured in seconds.
We then apply a soft logarithmic transformation, $8\cdot\ln(1+t/8)$, to compress the dynamic range of \emph{life-long} events while preserving high-resolution linear granularity for recent events.
This prevents aliasing in positional embeddings caused by extreme temporal gaps without sacrificing the precision of local event sequencing.
Second, to capture daily and weekly temporal cycles, we decompose each event timestamp into cyclical features for the hour of day, day of week, and day of month.
We embed these features using periodic functions similar to \citet{gorishniy2022embeddings}, but with periods fixed to the known calendar cycles rather than learned.
Calendar features are applied only to event-history entries, as cyclical patterns are less relevant for one-off life-long events where the log-seconds encoding already captures the relevant temporal signal.

\subsection{Model Architecture}
\label{sec:architecture}

\begin{figure}
    \centering
    \includegraphics[width=1\linewidth]{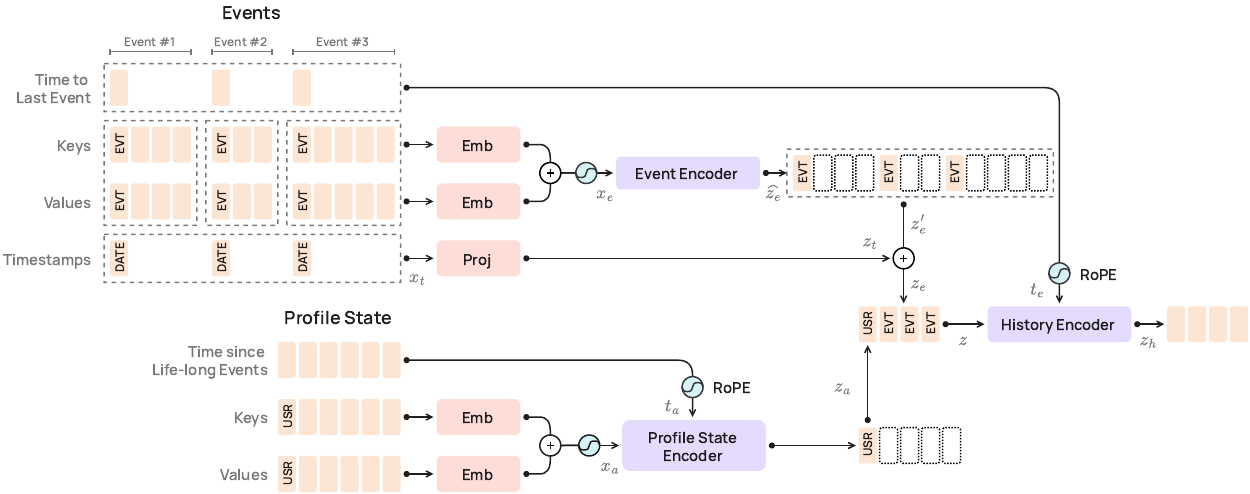}
    \caption{
        \textbf{PRAGMA backbone overview}.
        Each user record is represented as an ordered event history and profile state, where every field is decomposed into a semantic type (key), one or more values, and a temporal coordinate.
        Keys and values are embedded from a shared lookup table, and value tokens receive within-field positional embeddings.
        A \emph{Profile State Encoder} maps profile state $x_a$, with time since life-long events $t_a$ encoded via RoPE, into a \texttt{[USR]} embedding $z_a$, while an \emph{Event Encoder} independently maps the tokens of each event $x_e$ into a \texttt{[EVT]} embedding $z_e'$ and adds calendar features $z_t$.
        A \emph{History Encoder} then contextualises the sequence $z=[z_a:z_e]$ with time to the last event $t_e$ encoded via RoPE, producing a representation for a user record $z_h$.
    }
    \label{fig:method}
\end{figure}

PRAGMA is an encoder-only Transformer that takes an event history and contextual attributes as input and outputs dense record-level embeddings.
It is trained on a large-scale, diverse dataset with a masked modelling (MLM) objective that reconstructs masked input tokens.
Once pre-trained, it acts as a backbone for downstream adaptation with small-scale (2--4\,\% of the model's parameters) fine-tuning for a variety of tasks.
An overview of PRAGMA is shown in Figure~\ref{fig:method}.

PRAGMA is parametrised as a family of models with 10\,M, 100\,M, and 1\,B parameters, enabling selection according to operational budget and constraints.
The details of the architecture family are provided in Table~\ref{tab:model_family}.
All size variants use GELU activations~\citep{hendrycks2016gaussian}, pre-norm layer normalisation~\citep{xiong2020layer}, and dropout of 0.1~\citep{srivastava2014dropout}.

\begin{table}[t]
    \centering
    \begin{tabular}{lr rr rrr r}
        & & \multicolumn{2}{c}{\textbf{Width}} & \multicolumn{3}{c}{\textbf{Depth}} & \\
        \cmidrule(lr){3-4} \cmidrule(lr){5-7}
        \textbf{Model} & \textbf{Params} & \textbf{$d_{\mathrm{model}}$} & \textbf{$d_{\mathrm{ffn}}$} & \textbf{Profile} & \textbf{Event} & \textbf{History} & \textbf{Heads} \\
        \midrule
        PRAGMA-S &  10\,M & 192  & 768  & 1 & 5  & 2  & 3  \\
        PRAGMA-M & 100\,M & 512  & 2048 & 3 & 16 & 6  & 8  \\
        PRAGMA-L &   1\,B & 1024 & 4096 & 9 & 45 & 18 & 16 \\
    \end{tabular}
    \caption{
        \textbf{PRAGMA model family}.
        PRAGMA scales across three variants (10\,M, 100\,M, 1\,B parameters) by jointly increasing model width ($d_{\mathrm{model}}$, $d_{\mathrm{ffn}}$), depth of the profile-state, event, and history encoders, and the number of attention heads.
    }
    \label{tab:model_family}
\end{table}

The model consists of three main blocks: the Profile State Encoder, the Event Encoder, and the History Encoder.
First, the profile state tokens are processed by the Profile State Encoder.
Second, as with profile state, each event is encoded independently in the Event Encoder.
Finally, the outputs of the Profile State and Event Encoders are concatenated and encoded in the History Encoder to form an output.
Depending on the stage, the final output is used in an MLM head during pre-training, in a classification head during fine-tuning, or as-is in an embedding probe.

\subsubsection{Token Embedding}
\label{sec:token_embedding}
Profile state and event tokens are embedded identically.
For multi-valued fields (e.g., \texttt{Description}), the key token is replicated to match each of its values, yielding $n$ key--value pairs in total.
A single shared embedding table $E$ maps each key and value to a $d$-dimensional vector; the two embeddings are summed and augmented with static sine/cosine positional encodings (PosEmb)~\citep{vaswani2017attention}:
\begin{align}
    x = \text{PosEmb}\big(E(k) + E(v)\big), \quad x \inR^{n \times d}.
\end{align}
Positions index values \emph{within} a field, not across fields; e.g., the value \texttt{eur} of \texttt{Currency} receives position \texttt{0}, while the three value tokens \texttt{(met,\,al,\,plan)} of \texttt{Description} receive positions \texttt{(0,\,1,\,2)} (see Figure~\ref{fig:tokenisation}).
We denote user and event embeddings as $x_a \inR^{n_a \times d}$ and $x_e \inR^{n_e \times d}$, respectively.
Following common practice in encoder-only Transformers~\citep{devlin2019bert,dosovitskiy2021image}, a learnable \texttt{[USR]} (or \texttt{[EVT]}) token is prepended to each sequence (Figure~\ref{fig:method}).

\subsubsection{Profile State Encoder}
The Profile State Encoder is a bidirectional Transformer.
It takes as input the profile state tokens $x_a \inR^{n_a \times d}$ and corresponding temporal coordinates $t_a \inR^{n_a}$, where each entry holds the log-seconds since the corresponding life-long event (or $0$ for non-life-long pairs).
We use RoPE~\citep{su2024roformer} to encode $t_a$.
We disentangle this positional embedding from the value-level positional embedding discussed in~\S\ref{sec:token_embedding} to avoid the semantic and scale mismatch.
The output is a sequence of profile state embeddings $z_a\inR^{n_a \times d}$.
We pass the first element, which corresponds to the \texttt{[USR]} token, to the History Encoder; we refer to it as $z_a\inR^{1 \times d}$ for simplicity.

\subsubsection{Event Encoder}
\label{sec:event_encoder}
The Event Encoder is a bidirectional Transformer, similar to the Profile State Encoder.
It takes as input an event history $x_e = (x_{e, 1}, x_{e, 2}, \dots, x_{e, n_e})$, where each element contains a different number of token embeddings ($x_{e,i} \inR^{n_i \times d}$), and processes each event independently of all other events in the history.
The module outputs a token-level embedding sequence for each event, denoted $\widehat{z}_e$, which is used by the MLM head during pre-training.
Similar to the Profile State Encoder, we select the first token corresponding to the \texttt{[EVT]} token for each event as its aggregated representation $z_e' \inR^{n_e \times d}$.

The calendar features (hour of day, day of week, and day of month) $x_t\inR^{n_e\times 3}$ are converted to sine and cosine values and embedded with two MLP layers into $z_t\inR^{n_e \times d}$.
Next, the embedded calendar features are added to the Event Encoder output: $z_e = z_e' + z_t$.

\subsubsection{History Encoder}
The History Encoder is a bidirectional Transformer, similar to the other two encoders.
It takes as input the concatenated aggregated representations of profile state and the calendar-augmented events: $z=[z_a:z_e] \inR^{(1+n_e) \times d}$, as well as the corresponding temporal coordinate $t_e \inR^{1+n_e}$, where each entry holds the log-seconds to the most recent event in the history ($0$ for the $z_a$ position).
Similar to the Profile State Encoder, RoPE is used to encode positional information.
The output is a sequence of embeddings $z_h\inR^{(1+n_e)\times d}$, where $z_{h,0}$ corresponds to \texttt{[USR]} and $z_{h,1},\dots,z_{h,n_e}$ to the \texttt{[EVT]} tokens.
$z_h$ is used by the MLM head during pre-training and for downstream probes.

\subsubsection{Training}
\label{sec:training}
\paragraph*{Pre-training Objective.}
PRAGMA is pre-trained with an MLM objective following BERT~\citep{devlin2019bert}, in which a random subset of event input tokens is masked and the model reconstructs the original tokens.
For each masked token, the MLM head receives the concatenation of three $d$-dimensional vectors: the Event Encoder output at that token's position within $\widehat{z}_e$, providing local within-event context; the History Encoder output at the corresponding \texttt{[EVT]} position $z_{h,i}$, providing cross-event context; and the History Encoder output at the \texttt{[USR]} position $z_{h,0}$, providing user-level context.
This $3d$-dimensional representation is projected back to $d$ dimensions and matched against the embedding table to produce logits.
The training loss is cross-entropy with label smoothing~\citep{szegedy2016rethinking}.

\paragraph*{Masking Strategy.}
The masking strategy combines three sources: standard individual token-level masking (with 15\,\% probability), event-level masking (10\,\%) that requires the model to reconstruct an entire event, and semantic-type (key)-level masking (10\,\%) where all values of the selected keys are masked, training the model to predict values given context and a key.
During pre-training, a small fraction of selected positions are replaced with \texttt{[UNK]} rather than \texttt{[MASK]}.
Because \texttt{[UNK]} positions are excluded from the MLM objective, they receive no gradient and effectively act as a form of input dropout, training the model to recover original values under a stronger corruption scheme and reducing reliance on the presence of \texttt{[MASK]}, which does not occur at inference time.

\paragraph*{Downstream Adaptation.}
PRAGMA supports two modes of downstream adaptation.
In the \emph{embedding probe mode}, the record-level representation produced by the History Encoder is extracted as a frozen feature vector, and a lightweight linear probe is trained on top.
In the \emph{LoRA fine-tuning mode}, a small fraction (${\sim}$2--4\,\%) of model weights (the attention and feed-forward projections) are updated via Low-Rank Adaptation~\citep{hu2022lora}, keeping the pre-trained backbone mostly frozen and reducing the risk of catastrophic forgetting.

\subsection{Training Infrastructure}
\label{sec:train_infra}

Pre-training PRAGMA on 207\,B tokens spanning 24\,B user events introduces several engineering challenges.
The heterogeneous, table-structured nature of the data requires specialised storage, batching, and truncation strategies.
We describe each in turn below.

\paragraph*{Data Storage.}
The pre-training corpus is stored as a two-level structure: a \emph{user index} (an LMDB-backed key-value store mapping each user to their tokenised profile state and per-user token statistics) and a collection of \emph{event shards} (Parquet files partitioned by event count, so each file contains only users with the same number of events).
This layout allows workers to stream event shards independently and look up profile state on demand.

\paragraph*{Batching.}
Each training sample consists of a complete event history together with its associated profile state tokens.
Because event histories vary greatly in length, from a handful of events to thousands, na\"ive padding-based batching would waste the majority of compute on padding tokens.
Sharding records by event count avoids many random-access disk operations during loading and yields uniform-length event sequences within each batch, so the History Encoder operates on a rectangular tensor without ragged or padded dimensions.
We employ \emph{dynamic batching} with a fixed token budget that fits into GPU memory: records from the same shard are greedily packed until the budget is reached.

\paragraph*{Sequence Packing.}
Within a batch, individual events still vary in their number of tokens.
Rather than padding every event to the longest one, we pack all event tokens into a flat buffer and process them with a variable-length~(varlen) attention kernel~\citep{dao2022flashattention}, so tokens from different events do not attend to each other at this stage.
Together with shard-based batching, this eliminates padding overhead along both the event and token axes.
Compared to a padded baseline, sequence packing coupled with dynamic batching yields a $2$--$5{\times}$ throughput improvement, depending on the sequence length distribution in the dataset.

\paragraph*{Truncation.}
To bound memory consumption at a fixed context length, we apply two levels of truncation before packing.
At the \emph{event level}, each individual event is truncated to at most 24 tokens, affecting only 0.01\,\% of events.
At the \emph{profile state level}, the static profile state sequence is truncated to at most 200 tokens.
Users with zero events are discarded; users with more than 6{,}500 events are subsampled by retaining the most recent ones, preserving temporal recency.

\paragraph*{Pre-training Compute.}
The three model variants were trained with bf16 mixed precision and the Muon optimiser combined with AdamW~\citep{loshchilov2019adamw,jordan2024muon,liu2025muon}.
PRAGMA-S (10\,M parameters) and PRAGMA-M (100\,M) were trained on $16{\times}$\,NVIDIA H100 GPUs, and PRAGMA-L (1\,B) on $32{\times}$\,NVIDIA H100 GPUs.
The smallest variant converged in approximately 2~days, while the 100\,M and 1\,B models each required roughly 2~weeks of wall-clock time.

\section{Evaluation}
\label{sec:evaluation}
For commercial sensitivity reasons, we do not report absolute downstream metrics and instead express all results as relative changes with respect to a task-specific reference.
Throughout the paper, relative performance is computed as $(x / \text{baseline} - 1)\,\%$, where $x$ is the score of the evaluated method.

\subsection{Evaluation Protocol}
\label{sec:protocol}

We evaluate PRAGMA primarily via embedding probes and Low-Rank Adaptation (LoRA)~\citep{hu2022lora} fine-tuning on downstream tasks.

\subsubsection{Embedding Probing}
Embedding probing facilitates rapid iteration during experimentation before committing to LoRA fine-tuning, e.g., to gauge whether a new feature brings the expected gain, to select a checkpoint after a pre-training run for further evaluation, or to determine whether it is worth exploring a task as a downstream target at all.
The embeddings are extracted from the History Encoder output ($z_h$).

For our probing analysis, we evaluate the \texttt{[USR]} token, the final \texttt{[EVT]} token, and a combination of both, using a standard linear probe.
Given a downstream task with predefined train, validation, and test partitions, we first forward each record through the frozen encoder to obtain fixed-size representations and then train a linear probe (logistic or linear regression) on the training partition.
We observe that probe performance is robust to the choice of hyper-parameters, so fitting a probe typically takes a couple of minutes.
Since our architecture is inherently ``pre-norm'', the embeddings were standard-scaled prior to probe fitting.
We found that training the probe with the L-BFGS optimiser~\citep{liu1989limited} yielded the best results and converged quickly.

We note that while Gradient Boosted Decision Trees (GBDTs) perform well on lower-dimensional embeddings (e.g., $192$-d), the requirement for per-task hyper-parameter tuning and the increased time-to-fit make them less practical than linear probing for high-velocity model evaluation.

\subsubsection{Downstream Adaptation with LoRA}
To specialise the PRAGMA backbone for downstream tasks, we employ Low-Rank Adaptation (LoRA), which introduces a minimal parameter overhead of only 2--4\,\%.
In this setup, the pre-trained weights are fine-tuned for task-specific objectives to bridge the gap between general representation learning and downstream requirements.

We apply LoRA to the QKV projections and MLP layers within the encoder layers, following a common practice~\citep{hu2022lora,dettmers2023qlora}, and default to $\text{rank}=8$ with $\alpha=8$ across all experiments, but also sweep the rank across $\{4, 8, 16\}$ on smaller datasets.
LoRA fine-tuning with the Adam optimiser~\citep{kingma2015adam} typically takes one-eighth of the wall-clock time required for pre-training and converges in 12~hours to a few days, depending on the dataset size.

\subsubsection{Preparing Downstream Datasets}
For each downstream task, we obtain a unique identifier, which typically consists of a profile ID and an evaluation point.
Next, we gather the event history and profile attributes directly preceding the evaluation point.
We follow the pre-defined folds and splits for each downstream task.
The downstream dataset collection process mirrors that of the pre-training dataset.

\subsection{Downstream Tasks}
\label{sec:downstreams}

\paragraph*{Credit Scoring.}
The task is to assess credit risk for retail applications by predicting the probability of default within the first 12 months of use.
The downstream dataset spans multiple years and is diverse across records.
This task is cast as a binary classification problem with a minority class, and performance is measured with ROC-AUC and PR-AUC offline metrics.

\paragraph*{Communication Engagement.}
The task is to predict whether a user who abandoned a credit application mid-process will open a re-engagement communication.
This action serves as an upper-funnel proxy for resuming the application and eventually originating a loan.
A distinguishing aspect of this task is the severely limited sample size, requiring the model to capture nuanced event-level signals from minimal data.
This task is formulated as a binary classification problem, and the main offline metrics are ROC-AUC and PR-AUC.

\paragraph*{External Fraud.}
This task is a representative fraud detection use case formulated as a binary classification problem.
Performance is evaluated using precision and recall as the primary offline metrics.

\paragraph*{Product Recommendation.}
The task is to predict which products a user is likely to adopt in the near future, conditioned on receiving a specific communication (e.g., email or push notification).
A key challenge lies in modelling conversion propensity across multiple products simultaneously while accounting for the contextual influence of the communication.
The task is formulated as a multi-label classification problem, where the model outputs independent probabilities of conversion for each product in the portfolio.
Performance is evaluated using mean average precision (mAP) as the primary offline metric.

\paragraph*{Recurrent Transactions.}
This task focuses on predicting whether a given transaction corresponds to a recurring subscription that will repeat in the following month.
A key challenge lies in distinguishing true recurring patterns from irregular or one-off payments given limited historical signals.
The problem is formulated as a binary classification task, and performance is evaluated using macro-averaged $F_\text{1}$-score to account for class imbalance and ensure balanced performance across classes.

\paragraph*{Lifetime Value (LTV).}
The LTV task is to assess the probability of a user generating positive gross profit, and is formulated as a binary classification problem.
A distinguishing aspect of the LTV dataset is that users have shorter event histories, e.g., a couple of weeks, while the prediction horizon is typically 6 months or more.
The main offline metrics are ROC-AUC and PR-AUC.

\subsection{Main Results}
\label{sec:results}

The results presented in Table~\ref{tab:main_results} (left) demonstrate that PRAGMA-L consistently outperforms existing task-specific baselines across all evaluated domains, while the smaller variants remain competitive across nearly all combinations of tasks and metrics despite sharing most of their parameters across tasks.
For PRAGMA-L, the most striking improvements are observed in precision-recall metrics for high-impact tasks: PR-AUC increases by 130.2\,\% in Credit Scoring and 79.4\,\% in Communication Engagement, suggesting that PRAGMA is exceptionally effective at identifying low-frequency, high-value signals where traditional models struggle.
At the same scale, ROC-AUC gains are more tempered but remain substantial at +12.4\,\% and +20.4\,\% for the same tasks, respectively.
Although performance is more comparable on tasks like Lifetime Value and Recurrent Transactions, the overall trend confirms that PRAGMA provides a superior universal representation that matches or exceeds the performance of isolated, task-specific models.

\begin{table}[t]
\centering
\small
\setlength{\tabcolsep}{2.0pt}
\resizebox{\textwidth}{!}{%
\begin{tabular}{ll crrr | cr cr cr | cr}
\toprule
             &           & & \multicolumn{3}{c|}{\textbf{PRAGMA}} & \multicolumn{2}{c}{\textbf{S}} & \multicolumn{2}{c}{\textbf{M}} & \multicolumn{2}{c|}{\textbf{L}} & \multicolumn{2}{c}{\textbf{PRAGMA-M}} \\
\cmidrule(lr){4-6} \cmidrule(lr){7-8} \cmidrule(lr){9-10} \cmidrule(lr){11-12} \cmidrule(lr){13-14}
\textbf{Task} & \textbf{Metric} & \textbf{Baseline} & \multicolumn{1}{c}{\textbf{S}} & \multicolumn{1}{c}{\textbf{M}} & \multicolumn{1}{c|}{\textbf{L}} & \textbf{Emb.} & \multicolumn{1}{c}{\textbf{LoRA}} & \textbf{Emb.} & \multicolumn{1}{c}{\textbf{LoRA}} & \textbf{Emb.} & \multicolumn{1}{c|}{\textbf{LoRA}} & \textbf{Scratch} & \multicolumn{1}{c}{\textbf{LoRA}} \\
\midrule
Credit scoring & PR-AUC    & -- & +70.3\,\% & +98.0\,\% & +130.2\,\% & -- & +18.0\,\% & -- & +20.4\,\% & -- & +10.3\,\% & -- & +13.0\,\% \\
               & ROC-AUC   & -- &  +6.2\,\% & +10.1\,\% &  +12.4\,\% & -- &  +0.2\,\% & -- &  +2.4\,\% & -- &  +1.5\,\% & -- &  +1.6\,\% \\
\addlinespace
Comm.\ engagement & PR-AUC    & -- & +76.6\,\% & +76.7\,\% & +79.4\,\% & -- & +72.9\,\% & -- & +49.7\,\% & -- & +54.1\,\% & -- & +18.6\,\% \\
                  & ROC-AUC   & -- & +19.6\,\% & +17.4\,\% & +20.4\,\% & -- & +16.9\,\% & -- & +11.2\,\% & -- & +13.5\,\% & -- &  +5.0\,\% \\
\addlinespace
External fraud & Precision & -- &  +0.3\,\% & +12.3\,\% & +16.7\,\% & -- & +30.8\,\% & -- & +29.8\,\% & -- & +23.8\,\% & -- &  +8.0\,\% \\
               & Recall    & -- & +33.4\,\% & +66.4\,\% & +64.7\,\% & -- & +27.4\,\% & -- & +24.5\,\% & -- & +13.3\,\% & -- & +17.0\,\% \\
\addlinespace
Product rec.   & mAP           & -- & +10.6\,\% & +31.5\,\% & +40.5\,\% & -- & +57.2\,\% & -- & +68.4\,\% & -- & +68.1\,\% & -- & +10.3\,\% \\
\addlinespace
Recurrent txns & $F_\text{1}$ & -- &  +5.2\,\% &  +5.8\,\% &  +5.6\,\% & -- &  +4.5\,\% & -- &  +3.2\,\% & -- &  +2.3\,\% & -- &  +0.6\,\% \\
\addlinespace
Lifetime value & PR-AUC    & -- & -1.2\,\% &  +0.3\,\% &  +1.8\,\% & -- & +3.6\,\% & -- & +2.4\,\% & -- & +2.9\,\% & -- & +0.4\,\% \\
               & ROC-AUC   & -- & -0.8\,\% &  +0.9\,\% &  +2.6\,\% & -- & +4.7\,\% & -- & +3.4\,\% & -- & +3.9\,\% & -- & +0.3\,\% \\
\bottomrule
\end{tabular}
}
\caption{
    \textit{Left:} LoRA-adapted PRAGMA backbones are compared against task-specific baselines across scales (S/M/L).
    \textit{Middle:} Within each model size (S/M/L), LoRA is compared against embedding-only probes on the same backbone.
    \textit{Right:} At PRAGMA-M, LoRA fine-tuning of the pre-trained backbone is compared against training PRAGMA-M from scratch.
    All values are relative deltas $(\text{variant}/\text{ref}-1)\%$.
}
\label{tab:main_results}
\end{table}

\subsubsection{Effect of Model Scale}
\label{sec:scale_effect_result}
The left block of Table~\ref{tab:main_results} illustrates the performance impact of scaling the PRAGMA architecture from the Small~(S, 10\,M) variant to the Medium~(M, 100\,M) and Large~(L, 1\,B) variants.
We observe that scaling gains are highly task-dependent, with the most significant improvements concentrated in Credit Scoring: relative to the task-specific baseline, PR-AUC rises from +70.3\,\% at S to +98.0\,\% at M and +130.2\,\% at L, while ROC-AUC rises from +6.2\,\% to +10.1\,\% and +12.4\,\%, respectively.

Communication Engagement is non-monotonic in ROC-AUC, moving from +19.6\,\% at S to +17.4\,\% at M and +20.4\,\% at L relative to the task-specific baseline.
Recurrent Transactions remains nearly flat across scales ($+5.2/+5.8/+5.6\,\%$ $F_1$ for S/M/L), while Lifetime Value improves modestly from $-1.2/-0.8\,\%$ at S to $+1.8/+2.6\,\%$ at L in PR-AUC/ROC-AUC.
These results suggest that while increasing parameter count generally enhances predictive power, the Small model already provides a highly competitive representation for transactional and lifetime value predictions, offering a potential efficiency sweet spot for those specific production use cases.

\subsubsection{Effect of Pre-training}
\label{sec:pretraining_effect}

The right block of Table~\ref{tab:main_results} validates our approach, demonstrating that LoRA fine-tuning consistently matches or exceeds the performance of full-parameter training from scratch across all evaluated tasks.
The largest gains are observed in Communication Engagement, where LoRA achieves +18.6\,\% in PR-AUC and +5.0\,\% in ROC-AUC, suggesting that the pre-trained PRAGMA backbone captures rich, diverse event patterns that are difficult to learn when training a model from scratch on a single downstream task.
Credit Scoring follows a similar pattern, with LoRA yielding a +13.0\,\% improvement in PR-AUC and a +1.6\,\% lift in ROC-AUC.
Product Recommendation also benefits substantially, with a +10.3\,\% gain in mAP.
For Recurrent Transactions and Lifetime Value, the improvements are more modest (+0.6\,\% $F_1$, and +0.4\,\% / +0.3\,\% PR-AUC / ROC-AUC respectively), indicating that the scratch-trained baselines already capture most of the task-relevant structure for these objectives, and LoRA fine-tuning maintains parity without regression.
These findings are particularly significant for production environments, as they confirm that PRAGMA can consolidate multiple independent, high-maintenance models into a single shared system without sacrificing predictive accuracy, while maintaining a significantly smaller trainable parameter footprint.

\subsection{Additional Experiments and Ablations}
\label{sec:ablations}

\subsubsection{Effect of Low-Rank Adaptation}

\label{sec:emb_lora_results}
As shown in Table~\ref{tab:main_results} (middle), across all evaluated tasks and model scales, the LoRA-tuned variants consistently outperform the embedding-only baselines, demonstrating the efficacy of parameter-efficient fine-tuning in capturing task-specific nuances that fixed embeddings may miss.
The most substantial improvements are observed in Communication Engagement, where LoRA delivers a remarkable +72.9\,\% gain in PR-AUC for the Small model and maintains significant leads in the Medium and Large variants.
In Credit Scoring, we see a peak relative improvement of +20.4\,\% in PR-AUC for the Medium model, suggesting that LoRA layers are particularly effective at this scale for complex classification.
Gains in Recurrent Transactions and LTV are more modest, typically ranging from +2.3\,\% to +4.7\,\%.

\subsubsection{Robustness to Event Staleness}
\label{sec:staleness}

At serving time, PRAGMA reads a user's recent events from a shared event bus whose relaxed delivery guarantees can leave the latest events temporarily unobserved at scoring time.
For real-time external fraud detection, waiting for every event to propagate before scoring is impractical, making robustness to event staleness an important deployment requirement.
We evaluate this requirement by dropping recent events within a conservative window equal to the 99.9th percentile of observed propagation delays and comparing PRAGMA with the same model given the full event sequence.
For external fraud, neither precision nor recall drops by more than $0.2\,\%$ relative to the full sequence.
Credit scoring provides complementary evidence of stability, with neither ROC-AUC nor PR-AUC dropping by more than $0.9\,\%$.
These results indicate that PRAGMA remains reliable under the limited event staleness induced by eventual consistency, particularly for latency-sensitive fraud decisions.

\subsubsection{Effect of Profile State}
\label{sec:effect_profile_state}

Table~\ref{tab:pragma_s_results} isolates the contribution of the Profile State Encoder~(\S\ref{sec:architecture}) by comparing the full PRAGMA-S model against a variant that removes the profile-state branch entirely, relying solely on event-level representations.
The impact is strongly task-dependent.
Credit Scoring benefits substantially, with a +31.8\,\% relative gain in PR-AUC and +4.9\,\% in ROC-AUC.
The outsized PR-AUC improvement indicates that profile state is particularly valuable for identifying the minority default class, where static signals such as account tenure and onboarding characteristics provide discriminative context that event sequences alone cannot fully capture.
In contrast, Lifetime Value shows more moderate gains of +2.2\,\% in PR-AUC and +2.0\,\% in ROC-AUC, suggesting that gross-profit likelihood is largely inferable from transactional patterns over the prediction horizon.
Communication Engagement exhibits a slight PR-AUC regression ($-$3.0\,\%) alongside a marginal ROC-AUC gain (+1.3\,\%), indicating that re-engagement propensity is driven almost entirely by pre-drop-off event patterns rather than static user characteristics.
These results validate the two-branch design of PRAGMA: the dedicated Profile State Encoder adds significant value for tasks where static profile state is informative, while the architecture degrades gracefully when those signals are less relevant.

\begin{table}[t]
\centering
\begin{tabular}{llcr}
& & \multicolumn{2}{c}{\textbf{PRAGMA-S}} \\
\cmidrule(lr){3-4}
\textbf{Task} & \textbf{Metric} & \textbf{Event-only (ref.)} & \textbf{Full} \\
\midrule
External fraud           & Precision  & -- & +46.8\,\% \\
                         & Recall     & -- & +85.6\,\% \\
\addlinespace
Credit scoring           & PR-AUC     & -- & +31.8\,\% \\
                         & ROC-AUC    & -- & +4.9\,\% \\
\addlinespace
Product rec.   & mAP        & -- & +3.5\,\% \\
\addlinespace
Lifetime value           & PR-AUC     & -- & +2.2\,\% \\
                         & ROC-AUC    & -- & +2.0\,\% \\
\addlinespace
Recurrent txns   & $F_\text{1}$ & -- & +2.4\,\% \\
\addlinespace
Comm.\ engagement & PR-AUC     & -- & $-$3.0\,\% \\
                         & ROC-AUC    & -- & +1.3\,\% \\
\end{tabular}
\caption{
    \textbf{Profile state contributes substantially to tasks where static user characteristics are discriminative.}
    The relative performance is computed as ($\text{Full} / \text{Event-only} - 1$).
}
\label{tab:pragma_s_results}
\end{table}

\subsubsection{Communication Engagement (Uplift)}
\label{sec:credit_engagement_communications_uplift}

This task moves beyond conversion prediction to optimal treatment selection: the goal is to identify which messaging strategy best re-engages users with abandoned credit applications.
The dataset is smaller in scale than our other downstream benchmarks, yet large-scale pre-training proves decisive: the resulting model significantly outperforms a baseline trained on the limited in-domain data alone.
As an uplift task, it also offers a distinct evaluation angle: PRAGMA is used as a frozen feature extractor feeding a meta-learner rather than being fine-tuned, isolating representational quality in the absence of task-specific adaptation.

Concretely, we adopt a meta-learner framework~\citep{kunzel2019metalearners} to estimate heterogeneous treatment effects, requiring the model to capture complex interactions between pre-drop-off event signals, profile state, and treatment assignment.
Both PRAGMA and the baseline use the same meta-learner, differing only in the underlying representation.

Table~\ref{tab:uplift_results} summarises results using Area Under the Uplift Curve (AUUC) and SNIPS~\citep{swaminathan2015self}.
PRAGMA-L's ability to capture latent event-level patterns translates to highly effective treatment allocation, achieving a relative AUUC increase of 163.7\,\% over the internal baseline.

\begin{table}[t]
\centering
\begin{tabular}{llcr}
\textbf{Task}               & \textbf{Metric}   & \textbf{Baseline (ref.)} & \textbf{PRAGMA} \\
\midrule
Comm.\ engagement (uplift) & AUUC       & --                     & +163.7\,\%                \\
                           & SNIPS             & --                     &  +10.8\,\%                \\
\end{tabular}
\caption{
    \textbf{Performance comparison of PRAGMA-L against the internal uplift baseline using the same meta-learner framework.}
    The relative performance is computed as ($\text{PRAGMA-L} / \text{Baseline} - 1$).
    }
    \label{tab:uplift_results}
\end{table}

\subsubsection{Effect of a Pre-trained Text Encoder}
\label{sec:external_embeddings}

In the standard PRAGMA architecture, text values are learned jointly with all other tabular features via an embedding lookup table (see~\S\ref{sec:token_embedding}).
To prevent the model from underfitting sparse, noisy, or highly irregular financial text (e.g., truncated transaction descriptions), we investigate offloading text comprehension to a dedicated, pre-trained text embedding model, such as Nemotron-1B-v2~\citep{moreira2024nvretriever}.
This decoupled approach provides richer, out-of-the-box semantics and frees the primary Event Transformer~(\S\ref{sec:event_encoder}) to focus on cross-feature interactions.
While we do not use this as the default formulation in our generalised core architecture, we report on it as an optional extension that offers valuable domain-specific insights.

\begin{figure}[t]
    \centering
    \includegraphics[width=0.7\linewidth]{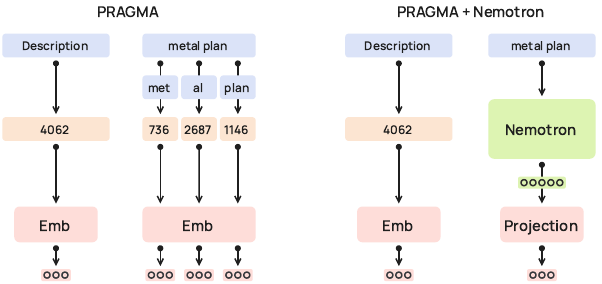}
    \caption{
        \textbf{Comparison of text embeddings with PRAGMA (left) and a version using pre-trained Nemotron-1B-v2 (right)}.
        Instead of our custom-trained BPE tokeniser and a trainable embedding lookup table, a pre-trained ``frozen'' Nemotron maps an entire text value to a single text embedding vector, which is then projected into the Transformer's base dimension with a trainable projection.
    }
    \label{fig:external_emb_arch}
\end{figure}

\paragraph*{Implementation Details.}
The addition of a pre-trained text encoder involves multiple structural changes to the PRAGMA architecture.
First, for semantic types (keys) whose values are normally encoded using a custom-trained BPE tokeniser and a trainable embedding lookup table, we instead use the frozen pre-trained model to map the complete text string to a single vector, which is then adapted via a one-layer trainable projection (see Figure~\ref{fig:external_emb_arch}).
Second, instead of reconstructing exact token labels for these text fields during MLM optimisation (see~\S\ref{sec:training}), we train PRAGMA to reconstruct the continuous text embedding produced by the pre-trained text encoder with Mean Squared Error (MSE).

\paragraph*{Results \& Discussion.}
The results are shown in Table~\ref{tab:pretrained_text_embedding}.
Downstream effects track how much label-relevant signal sits in free text versus categorical and behavioural structure.
Credit Scoring shows the clearest upside, with +16.1\,\% relative PR-AUC and +2.8\,\% ROC-AUC under Nemotron.
Product Recommendation instead loses ground: mAP drops by 6.4\,\% relative, plausibly because sparse text adds little beyond what the structural channels already encode.
External Fraud moves modestly and in opposite directions on precision (+3.8\,\%) versus recall ($-$0.7\,\%), while LTV and Recurrent Transactions stay near-flat on the reported metrics.
Because this variant also increases PRAGMA-M training latency by about 18\,\%, we keep it as an opt-in module for text-heavy tasks rather than baking it into the default architecture.

\begin{table}[t]
\centering
\begin{tabular}{llcr}
& & \multicolumn{2}{c}{\textbf{PRAGMA-M}} \\
\cmidrule(lr){3-4}
\textbf{Task}            & \textbf{Metric} & \textbf{ref.} & \textbf{+Nemotron} \\
\midrule
Credit scoring           & PR-AUC          & --                     & +16.1\,\%                   \\
                         & ROC-AUC         & --                     &  +2.8\,\%                   \\
\addlinespace
Recurrent txns   & $F_\text{1}$    & --                     &  +0.1\,\%                   \\
\addlinespace
Lifetime value           & PR-AUC          & --                     &  +0.8\,\%                   \\
                         & ROC-AUC         & --                     &  +0.6\,\%                   \\
\addlinespace
External fraud           & Precision       & --                     &  +3.8\,\%                   \\
                         & Recall          & --                     &  $-$0.7\,\%                   \\
\addlinespace
Product rec.   & mAP             & --                     &  $-$6.4\,\%                  \\
\end{tabular}
\caption{
    \textbf{Impact of pre-trained text embeddings on downstream tasks is concentrated in text-heavy domains.}
    The performance is estimated relative to a LoRA-tuned PRAGMA-M.
}
\label{tab:pretrained_text_embedding}
\end{table}

\subsubsection{Limitations in Highly Relational Tasks: Anti-Money Laundering}
\label{sec:aml}
We formulate Anti-Money Laundering (AML) as a binary classification task.
As shown in Table~\ref{tab:aml_results}, this is a setting where PRAGMA significantly underperforms the production baseline.

We attribute this performance gap to two primary factors.
First, the downstream AML dataset is sufficiently large for the baseline model to learn robust task-specific representations without requiring foundation-level pre-training.
Second, and more critically, AML detection is inherently relational: the baseline leverages cross-record features that capture network-level signals.
Because PRAGMA processes event histories in isolation, the resulting embeddings do not inherently capture the cross-record dependency structures crucial for this task.

Performance is evaluated primarily using $F_\text{0.5}$, as it emphasises precision while still accounting for recall.
PRAGMA suffers a 47.1\,\% drop in $F_\text{0.5}$ compared to the network-aware baseline, demonstrating that isolated record-level representations may be insufficient for this highly relational domain.
Addressing this limitation remains a key direction for future work.

\begin{table}[t]
\centering
\begin{tabular}{llcr}
\textbf{Task}               & \textbf{Metric}   & \textbf{Baseline (ref.)} & \textbf{PRAGMA} \\
\midrule
Anti-money laundering       & $F_\text{0.5}$    & --                     & $-$47.1\,\%        \\
\end{tabular}
\caption{
    \textbf{Performance comparison of PRAGMA against baseline for Anti-Money Laundering.}
    The relative performance is computed as ($\text{PRAGMA} / \text{Baseline} - 1$) using a linear probe on PRAGMA-L embeddings.
    }
    \label{tab:aml_results}
\end{table}

\section{Related Work}
\label{sec:related-work}

\subsection{Transformer}
\label{sec:rw-transformer}

The landscape of sequence modelling was fundamentally reshaped by the introduction of the Transformer architecture~\citep{vaswani2017attention}, which dispensed with recurrent layers in favour of a parallelisable self-attention mechanism.
Following this, the field branched out into encoder-only models like BERT~\citep{devlin2019bert}, optimised for discriminative tasks, and decoder-only architectures like GPT-3~\citep{brown2020language}, which catalysed the current generative AI era through massive scaling and emergent in-context learning.
Subsequent research has extended the architecture's reach via the Vision Transformer (ViT)~\citep{dosovitskiy2021image} for visual perception and the T5 framework~\citep{raffel2020exploring} for unified text-to-text processing.
Recent advancements have prioritised computational efficiency and multimodality, notably through hardware-aware optimisations like FlashAttention~\citep{dao2022flashattention} and the adoption of Mixture-of-Experts (MoE)~\citep{fedus2022switch} in models like Mixtral $8{\times}7$B~\citep{jiang2024mixtral}.
In the current paradigm, models such as Gemini~1.5~\citep{gemini2024gemini} and GPT-4o~\citep{hurst2024gpt} have moved beyond compositional architectures to native multimodality, enabling seamless reasoning across diverse data streams.

In this landscape, PRAGMA should be understood as an encoder foundation model for heterogeneous tabular event streams. Although motivated by financial transactions, it extends naturally to any domain where entities accumulate irregular, multi-field records over time.
It inherits the scalability and bidirectional contextualisation of encoder-only Transformers, adapting them to heterogeneous fields, explicit time signals, and reusable record-level representations.

\subsection{Masked Modelling}
\label{sec:rw-masked-modelling}

Parallel to the scaling of generative decoders, masked modelling established a dominant paradigm for self-supervised representation learning.
This was pioneered by BERT~\citep{devlin2019bert}, which utilised a Masked Language Modelling (MLM) objective to capture bidirectional context, a technique further refined by RoBERTa~\citep{liu2019roberta} through dynamic masking and optimised training recipes.
The success of MLM was later translated to the vision domain via Masked Image Modelling (MIM), with BEiT~\citep{bao2021beit} and Masked Autoencoders (MAE)~\citep{he2022masked} demonstrating that reconstructing obscured image patches forces the model to learn holistic structural representations.
Recent trends have pursued cross-modal unification, as seen in Data2Vec~\citep{baevski2022data2vec}, and a shift from raw signal reconstruction to latent feature prediction, exemplified by the Joint-Embedding Predictive Architecture (I-JEPA)~\citep{assran2023self}.

PRAGMA is directly inspired by this line of work, but extends masked modelling from text and images to heterogeneous financial records.
Our objective masks individual tokens, whole events, and semantic types, encouraging the reconstruction of partially observed events and the learning of transferable representations from full transaction histories.

\subsection{Transformers for Tabular Data}
\label{sec:rw-tabular}

While Gradient Boosted Decision Trees (GBDTs) have historically dominated structured data, the Transformer has spurred a new class of ``Tabular Deep Learning'' architectures.
Early entries like TabTransformer~\citep{huang2020tabtransformer} and FT-Transformer~\citep{gorishniy2021revisiting} focused on modelling inter-feature dependencies through self-attention, demonstrating performance parity with GBDTs on high-dimensional datasets.
This was improved by SAINT~\citep{somepalli2021saint}, which introduced a dual-attention mechanism for both feature and row interactions, and Trompt~\citep{chen2023trompt}, which proposed prompt-tuning to disentangle intrinsic table properties from sample variations.
A paradigm shift occurred with TabPFN~\citep{hollmann2023tabpfn}, a foundation model pre-trained on synthetic data to approximate Bayesian inference.
Leveraging in-context learning, TabPFN generates predictions via a single forward pass, eliminating the need for iterative training.
While the original model was restricted to 1{,}000 samples, TabPFN-v2 and TabPFN-v2.5~\citep{hollmann2025accurate,grinsztajn2025tabpfn}~scaled the architecture to handle 100{,}000 samples and real-world complexities, providing native support for categorical features, missing values, and outliers.
Most recently, Mitra~\citep{zhang2025mitra} has adopted the dual-attention mechanism of SAINT but follows the foundation model paradigm of TabPFN by being pre-trained exclusively on a massive mixture of synthetic priors.

PRAGMA is related in spirit to tabular Transformers because it preserves field identity and models cross-field interactions with attention, but unlike TabTransformer, FT-Transformer, and SAINT, it does not operate on a single row with a fixed schema.
Compared with TabPFN-style tabular foundation models trained on synthetic supervised tasks, PRAGMA is pre-trained with self-supervision on real financial ledgers and models variable-length user histories of heterogeneous events with a hierarchical encoder.

\subsection{Modelling for Recommender Systems}
\label{sec:rw-recsys}

Sequential recommendation models share structural similarities with transaction modelling, as both process ordered event sequences with rich side information.
Transformer-based recommenders treat user interaction histories as token sequences: SASRec~\citep{kang2018self} replaced recurrence with self-attention to capture long-range dependencies, and BERT4Rec~\citep{sun2019bert4rec} demonstrated that bidirectional context via masked item prediction yields more robust representations.
The field later converged with the LLM paradigm: P5~\citep{geng2022recommendation} cast diverse recommendation tasks into a unified text-to-text framework built on~T5, while TALLRec~\citep{bao2023tallrec} introduced instruction tuning to align general-purpose LLMs with recommendation logic.

More recent industrial work has shifted from modelling only positive interactions to encoding richer event streams.
Generative Recommenders~\citep{zhai2024actions} interleave item and action tokens in a causal sequence, scaling to trillions of parameters with power-law quality gains.
ARGUS~\citep{khrylchenko2025scaling} decomposes autoregressive learning into feedback and next-item prediction, scaling recommender Transformers to one~billion parameters.
The TransAct line of work~\citep{xia2023transact,xia2025transact} embeds each user action as a composite of content, action type, and context for CTR prediction, and extends to lifelong action sequences.

PRAGMA is close to this literature in its use of ordered event histories and self-supervised pre-training.
Unlike recommendation models that often reduce each interaction to an item token, PRAGMA models richer financial events with typed fields, amounts, free text, and temporal coordinates, and is adapted to a broader set of banking tasks beyond ranking.

\subsection{Foundation Models for Finance}
\label{sec:rw-finance}

The paradigm of financial foundation models has rapidly matured from specialised text encoders to comprehensive reasoning engines that integrate diverse data modalities.
This evolution began with FinBERT~\citep{yang2020finbert}, which adapted the encoder-only architecture to financial corpora, establishing a rigorous baseline for discriminative tasks like sentiment analysis and ESG classification.
The field shifted toward massive generative scale with BloombergGPT~\citep{wu2023bloomberggpt}, which demonstrated that interleaving proprietary financial datasets with general web corpora yields superior performance on domain-specific benchmarks.
To address the accessibility barriers of such massive models, FinGPT~\citep{yang2023fingpt} introduced a data-centric, lightweight adaptation framework, democratising access to financial LLMs via efficient LoRA fine-tuning~\citep{hu2022lora} of open-source models.
Most recently, research has transcended textual boundaries to address the structured nature of market data; models like Time-LLM~\citep{jin2024timellm} and Chronos~\citep{ansari2024chronos} treat numerical time series as token sequences, enabling Transformers to perform zero-shot forecasting.

Extending this structural shift to consumer finance, recent foundation models are now being trained directly on massive-scale user transaction ledgers.
For instance, nuFormer~\citep{braithwaite2025your} pre-trains a decoder-only Transformer solely on tokenised transaction sequences.
During task-specific fine-tuning, nuFormer adds a DCNv2 tabular network~\citep{wang2021dcn} to blend the resulting embeddings with the downstream baseline's existing hand-crafted feature set.
Concurrently, TransactionGPT~\citep{dou2025transactiongpt} introduces a specialised 3D-Transformer that combines field-level fusion with autoregressive modelling over transaction histories.
It is trained and evaluated separately in three payment settings: joint transaction generation and anomaly detection, next-restaurant prediction, and next-merchant-category prediction.

PRAGMA differs from text-centric financial foundation models such as FinBERT, BloombergGPT, and FinGPT, which primarily operate on financial language, and from Time-LLM or Chronos, which tokenise numerical time series for forecasting.
It is closer to nuFormer and TransactionGPT, but jointly pre-trains an encoder backbone with dedicated branches for multi-source banking events and task-agnostic profile state, then adapts the same backbone across six downstream tasks.

\section{Conclusion}
\label{sec:conclusion}

We presented PRAGMA, a family of encoder-style foundation models for multi-source banking user histories.
PRAGMA combines a key--value--time tokenisation scheme with two encoder branches, one for profile state and one for events.
Their outputs are fused by a history encoder, and the model is pre-trained with masked modelling on large-scale, heterogeneous financial records.
Across diverse downstream tasks (credit scoring, fraud detection, communication engagement, product recommendation, recurrent transaction detection, lifetime value prediction, and more), a single pre-trained backbone achieves superior performance directly from raw banking event sequences, providing a general-purpose representation layer for financial applications.

Our experiments reveal several practical insights.
LoRA fine-tuning consistently matches or exceeds full training from scratch while updating only a small fraction of parameters, confirming that the pre-trained representations transfer effectively across tasks.
Scaling from 10\,M to 1\,B parameters yields large gains on harder tasks such as credit scoring, while smaller models already provide competitive representations for tasks such as lifetime value prediction, offering a practical efficiency trade-off.
The dedicated profile state encoder proves particularly valuable for tasks where static contextual attributes are informative, such as credit scoring and fraud detection, while the architecture degrades gracefully when those signals are less relevant.
We also find that integrating a pre-trained text encoder improves performance in text-dense domains but adds training overhead that is not justified for text-sparse tasks.
Finally, the AML case study highlights a clear limitation: tasks that depend on cross-record relational structure remain out of reach for a model that processes event histories in isolation.

These results suggest that multi-source banking event sequences admit transferable representations in much the same way as text and vision, despite their heterogeneous structure, irregular timing, and operational constraints.
Extending the model to capture cross-record interactions for relational tasks such as anti-money laundering is a promising direction for future work.

\subsubsection*{Acknowledgments}

We thank
Dmitry Mittov,
Ian Iakobsen,
Aleksandr Pushin,
Muhammad Anas,
Viacheslav Karpov,
Nathalie Skrzypek,
Leyla Sultanova,
Francisco Sanz Estevez,
Nikita Kravchuk,
Tadas Krisciunas,
Amey Baokar,
Hanna Danilovich,
Jyoti Prakash Bal,
Vitalii Radchenko,
Kade Main,
Nic Hatia,
and other Revoluters for their contributions to this work.

\bibliography{references}
\bibliographystyle{tmlr-firstauthor}

\end{document}